# The Multiscenario Multienvironment BioSecure Multimodal Database (BMDB)

Javier Ortega-Garcia, *Senior Member*, *IEEE*, Julian Fierrez, *Member*, *IEEE*,
Fernando Alonso-Fernandez, *Member*, *IEEE*, Javier Galbally, *Student Member*, *IEEE*,
Manuel R. Freire, Joaquin Gonzalez-Rodriguez, *Member*, *IEEE*,
Carmen Garcia-Mateo, *Member*, *IEEE*, Jose-Luis Alba-Castro, *Senior Member*, *IEEE*,
Elisardo Gonzalez-Agulla, Enrique Otero-Muras, Sonia Garcia-Salicetti, *Member*, *IEEE*,
Lorene Allano, Bao Ly-Van, Bernadette Dorizzi, Josef Kittler, Thirimachos Bourlai,
Norman Poh, *Member*, *IEEE*, Farzin Deravi, *Member Computer Society*, *IEEE*,
Ming W.R. Ng, *Member*, *IEEE*, Michael Fairhurst, Jean Hennebert, *Member*, *IEEE*,
Andreas Humm, *Member*, *IEEE*, Massimo Tistarelli, *Senior Member*, *IEEE*, Linda Brodo,
Jonas Richiardi, *Member*, *IEEE*, Andrzej Drygajlo, *Member*, *IEEE*, Harald Ganster, Federico M. Sukno,
Sri-Kaushik Pavani, Alejandro Frangi, *Senior Member*, *IEEE*,
Lale Akarun, *Senior Member*, *IEEE*, and Arman Savran

**Abstract**—A new multimodal biometric database designed and acquired within the framework of the European BioSecure Network of Excellence is presented. It is comprised of more than 600 individuals acquired simultaneously in three scenarios: 1) over the Internet, 2) in an office environment with desktop PC, and 3) in indoor/outdoor environments with mobile portable hardware. The three scenarios include a common part of audio/video data. Also, signature and fingerprint data have been acquired both with desktop PC and mobile portable hardware. Additionally, hand and iris data were acquired in the second scenario using desktop PC. Acquisition has been conducted by 11 European institutions. Additional features of the BioSecure Multimodal Database (BMDB) are: two acquisition sessions, several sensors in certain modalities, balanced gender and age distributions, multimodal realistic scenarios with simple and quick tasks per modality, cross-European diversity, availability of demographic data, and compatibility with other multimodal databases. The novel acquisition conditions of the BMDB allow us to perform new challenging research and evaluation of either monomodal or multimodal biometric systems, as in the recent BioSecure Multimodal Evaluation campaign. A description of this campaign including baseline results of individual modalities from the new database is also given. The database is expected to be available for research purposes through the BioSecure Association during 2008.

---

## 1 INTRODUCTION

**B**IOMETRICS is defined as the use of anatomical and/or behavioral traits of individuals for recognition purposes [1]. Advances in the field would not be possible without suitable biometric data in order to assess systems' performance through benchmarks and evaluations. In past years, it was common to evaluate biometric products on small custom or proprietary data sets [2], and therefore, experiments were not repeatable and a comparative assessment could not be accomplished. As biometric systems are being deployed, joint efforts have been conducted to perform common experimental protocols and technology benchmarks. Several evaluation procedures [3], databases, and competitions have been established in recent years, e.g., the NIST Speaker Recognition Evaluations [4], the FERET and FRVT Face Recognition Evaluations [5], the series of Fingerprint Verification Competitions (FVC) [6], the Iris Challenge Evaluation (ICE) [7], or the Signature Verification Competition (SVC) [8]. Biometric data gathered for these competitions along with experimental protocols are made publicly available, and in many cases, they have become a *de facto* standard so that data and protocols utilized in these competitions are used by the biometric community to report subsequent research work.

However, evaluation of biometric systems has been done to date in a fragmented way, modality by modality, without any common framework. Some efforts in the past have been made regarding multimodal research and applications. Several multimodal databases are available today, typically

---

- *J. Fierrez is with the ATVS/Biometric Recognition Group, Escuela Politecnica Superior, Univ. Autonoma de Madrid, Avda. Francisco Tomas y Valiente 11, Campus de Cantoblanco, 28049 Madrid, Spain. He is the corresponding author for this paper. E-mail: julian.fierrez@uam.es.*

as a result of collaborative European or national projects, but most of them include just a few modalities. Biometric database collection is a challenging task [2]. The desirability of large databases with the presence of variability (multi-session, multisensor, multiscenario, etc.) makes the acquisition a time and resource-consuming process, which, in the case of multimodal databases, requires important additional efforts. A number of preacquisition and post-acquisition tasks are also needed: training of acquisition operators, recruitment of donors, supervision of acquired data, annotation, error correction, labeling, documentation, etc. Furthermore, legal and operational issues surround the donor's consent and storage and distribution process. Usually, a license agreement has to be signed between the distributor and the licensee, and the size of new databases (from gigabytes to terabytes) complicates their distribution.

This paper presents the recently acquired BioSecure Multimodal Database, which was collected between November 2006 and June 2007. An important integrative effort has been done in the design and collection of this database, involving 11 European institutions of the BioSecure Network of Excellence [9]. The new database includes new features not present in existing databases. More than 600 individuals have been acquired simultaneously in three different scenarios (over the Internet, in an office environment with a desktop PC, and in indoor/outdoor environments using mobile devices) over two acquisition sessions and with different sensors for certain modalities. New challenging acquisition conditions that will allow the evaluation of realistic scenarios are included, such as Internet transactions using face and voice acquired with commercial Webcams, or fingerprint and handwritten signature on modern mobile platforms. An example of such an evaluation is the recently conducted BioSecure Multimodal Evaluation Campaign (BMEC) [10], where several tasks were defined using data extracted from the BioSecure Multimodal Database. Being the largest publicly available multimodal database, it will allow novel research like: subpopulation effects, only attainable by having a large population; large scale multi-scenario/multisensor interoperability tests; etc. Furthermore, the width of the database gives us the ability to have very tight confidence intervals on results.

The contributions of the present paper can be summarized as: 1) an overview of existing resources for research on multimodal biometrics and 2) a detailed description of the most comprehensive multimodal biometric database available to date, including aspects such as: relation with other existing databases, acquisition design, postprocessing tasks, legal issues regarding distribution of biometric data, and a core experimental framework with baseline results that can be used as a reference when using the database. From these baseline results, some novel experimental findings can also be observed in topics of practical importance such as biometric recognition using mismatched devices, and multiple acquisition environments.

The rest of this paper is organized as follows: Section 2 summarizes previous work done in the domain of multimodal biometric database design and acquisition. The BioSecure Multimodal Database (BMDB) is described in Section 3, and its compatibility with existing databases is given in Section 4. A brief description of the BioSecure Multimodal Evaluation Campaign, together with baseline results of individual modalities from the database, is provided in Section 5. Some research directions that can be followed with the new databases are pointed out in Section 6, and finally, legal and distribution issues are discussed in Section 7.

## 2 RELATED WORKS

Multimodal biometric research was first supported by "chimeric" multimodal databases containing synthetic subjects [11]. These synthetic subjects were constructed by combining biometric data from different databases (e.g., fingerprint images from one subject in one database, face images from another subject in another database, etc.) to produce data sets which do not represent real multimodal samples. As multibiometric data may be necessarily correlated [12], the use of chimeric databases should be avoided. But collecting multimodal databases has important issues: More acquisition time is generally required, subjects may react negatively to the longer acquisition sessions needed, the size of the database and acquisition cost are considerably higher, and in general, the management of such a task is exponentially more complex. Fortunately, in recent years, efforts have been directed toward collecting real multimodal databases involving various biometric traits and hundreds of users.

First, efforts were focused on the acquisition of mono-modal or bimodal databases (one or two biometric traits sensed), e.g., the **MCYT** database [20], including signatures and fingerprint images of 330 subjects; the **M2TVS** [24], **XM2VTS** [23], and **BANCA** [21] databases, with face and voice data of 37, 295, and 208 subjects, respectively; or the **FRGC** database [19], which includes 2D and 3D face images of 741 subjects; and the **BT-DAVID** database [25], with audiovisual data from 124 individuals.

There are also several multimodal biometric databases with multiple traits available today, or in the process of completion, mainly as a result of collaborative national or international projects. Some examples include: the **Smart-Kom** [22], **M3** [18], and **MBioID** [17] databases, and the following ones (which we present in some more detail because of their relation to the BioSecure multimodal database):

- **BiosecurID** database [13]. The BiosecurID database was collected in six Spanish institutions in the framework of the BiosecurID project funded by the Spanish Ministry of Education and Science. It has been collected in an office-like uncontrolled environment (in order to simulate a realistic scenario), and was designed to comply with the following characteristics: 400 subjects, eight different traits (speech, iris, face still and talking face, signature, handwriting, fingerprint, and hand and keystroking), and four acquisition sessions distributed in a four-month time span.
- **BioSec** database [14]. BioSec was an Integrated Project (IP) of the Sixth European Framework Programme [26] which involved over 20 partners from nine European countries. One of the activities within BioSec was the acquisition of a multimodal database. This database was acquired at four different European

TABLE 1
Most Relevant Features of Existing Multimodal Databases

| | Year | Ref. | Users | Sessions | Traits | 2Fa | 3Fa | Fp | Ha | Hw | Ir | Ks | Sg | Sp |
|---|---|---|---|---|---|---|---|---|---|---|---|---|---|---|
| **BioSecure** | 2008 | - | 971 (DS1, Internet scenario) | 2 | 2 | X | | | | | | | | X |
| | | | 667 (DS2, Desktop scenario) | 2 | 6 | X | | X | X | | X | | X | X |
| | | | 713 (DS3, Mobile scenario) | 2 | 4 | X | | X | | | | | X | X |
| **BiosecurID** | 2007 | [13] | 400 | 4 | 8 | X | | X | X | X | X | X | X | X |
| **BioSec** | 2007 | [14] | 250 | 4 | 4 | X | | X | | | X | | | X |
| **MyIDEA** | 2005 | [15] | 104 (approx.) | 3 | 6 | X | | X | X | X | | | X | X |
| **BIOMET** | 2003 | [16] | 91 | 3 | 6 | X | X | X | X | | | | X | X |
| **MBioID** | 2007 | [17] | 120 (approx.) | 2 | 6 | X | X | X | | | X | | X | X |
| **M3** | 2006 | [18] | 32 | 3 | 3 | X | | X | | | | | | X |
| **FRGC** | 2006 | [19] | 741 | variable | 2 | X | X | | | | | | | |
| **MCYT** | 2003 | [20] | 330 | 1 | 2 | | | X | | | | | X | |
| **BANCA** | 2003 | [21] | 208 | 12 | 2 | X | | | | | | | | X |
| **Smartkom** | 2002 | [22] | 96 | variable | 4 | X | X | | | | | | X | X |
| **XM2VTS** | 1999 | [23] | 295 | 4 | 2 | X | | | | | | | | X |
| **M2VTS** | 1998 | [24] | 37 | 5 | 2 | X | | | | | | | | X |
| **BT-DAVID** | 1999 | [25] | 124 | 5 | 2 | X | | | | | | | | X |

*The nomenclature is as follows: 2Fa stands for Face 2D, 3Fa for Face 3D, Fp for Fingerprint, Ha for Hand, Hw for Handwriting, Ir for Iris, Ks for Keystrokes, Sg for Handwritten Signature, and Sp for Speech.*

sites and includes face, speech (both with a Webcam and a headset), fingerprint (with three different sensors), and iris recordings. The baseline corpus [14] is comprised of 200 subjects with two acquisition sessions per subject. The extended version of the BioSec database is comprised of 250 subjects with four sessions per subject (about one month between sessions). A subset of this database was used in the last International Fingerprint Verification Competition [6] held in 2006.

- **MyIDEA** database [15], which includes face, audio, fingerprint, signature, handwriting, and hand geometry of 104 subjects. Synchronized face-voice and handwriting-voice were also acquired. Sensors of different quality and various scenarios with different levels of control were considered in the acquisition.
- **BIOMET** database [16], which offers five different modalities: audio, face images (2D and 3D), hand images, fingerprint (with an optical and a capacitive sensor), and signature. The database consists of three different acquisition sessions (with eight months between the first and the third) and is comprised of 91 subjects who completed the three sessions.

In Table 1, the most relevant features of the existing multimodal databases are summarized. The current paper presents the recently acquired BioSecure Multimodal Database. It is designed to comply with several characteristics that, as can be observed in Table 1, make it unique, namely: hundreds of users and several biometric modalities acquired under several scenarios.

## 3 THE BIOSECURE MULTIMODAL DATABASE

### 3.1 General Description

The acquisition of the BMDB was jointly conducted by 11 European institutions participating in the BioSecure Network of Excellence [9], see Table 2. The institution in charge of coordinating the acquisition process was the Universidad Politecnica de Madrid (UPM), from Spain, through the ATVS/Biometric Recognition Group (marked with (*) in Table 2). BMDB is comprised of three different data sets, with an institution in charge of coordinating the acquisition of each data set (marked with ($\triangle$) in Table 2). The three data sets are:

- **Data Set 1 (DS1)**, acquired over the **Internet** under unsupervised conditions (i.e., connecting to an

TABLE 2
Institutions Participating in the Acquisition,
Including Involvement in the Three Acquired Data Sets

| INSTITUTION | COUNTRY | DS1 | DS2 | DS3 |
|---|---|---|---|---|
| Universidad de Vigo | Spain | 101 ($\triangle$) | 101 | - |
| Bogazici University | Turkey | 98 | - | 95 |
| Ecole Polytechnique Federale de Lausanne | Switzerland | 77 | - | 48 |
| Groupe des Ecoles des Telecommunications | France | 117 | 117 | 117 ($\triangle$) |
| Joanneum Research Graz | Austria | 70 | 70 | - |
| University of Fribourg | Switzerland | 70 | - | 70 |
| University of Kent | UK | 80 | 80 | 80 |
| University of Surrey | UK | 104 | 104 | 104 |
| University of Sassari | Italy | 55 | 55 | - |
| Pompeu Fabra University | Spain | 59 | - | 59 |
| Universidad Politecnica de Madrid (*) | Spain | 140 | 140 ($\triangle$) | 140 |
| | | 971 | 667 | 713 |

*For each data set, there was an institution—marked with ($\triangle$)—in charge of coordinating its acquisition. (*) The overall acquisition process was coordinated by the Universidad Politecnica de Madrid (UPM) through the ATVS/Biometric Recognition Group. The ATVS group, formerly at the Universidad Politecnica de Madrid (UPM), is currently at Universidad Autonoma de Madrid (UAM). The Groupe des Ecoles des Telecommunications (GET) has changed its name to Institut TELECOM.*

TABLE 3
Statistics of the Three Data Sets
of the BioSecure Multimodal Database

| | DS1 | DS2 | DS3 |
|---|---|---|---|
| Age distribution (18-25/25-35/35-50/>50) | 43% / 25% / 19% / 23% | 41% / 21% / 21% / 17% | 42% / 25% / 17% / 16% |
| Gender distribution (male / female) | 58% / 42% | 56% / 44% | 56% / 44% |
| Handedness (righthanded / lefthanded) | 94% / 6% | 94% / 6% | 94% / 6% |
| Manual workers (yes / no) | 2% / 98% | 3% / 97% | 2% / 98% |
| Vision aids (glasses, contact lenses / none) | 42% / 58% | 42% / 58% | 43% / 57% |

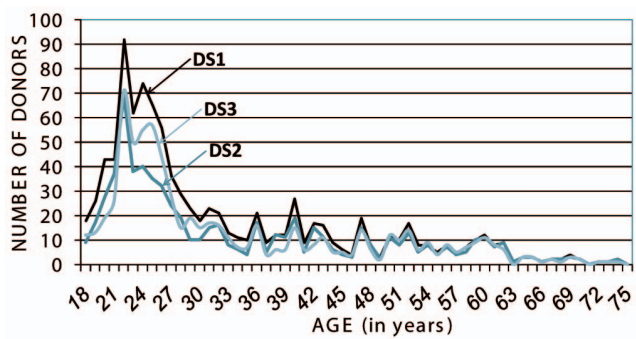

Fig. 1. Age distribution of the three data sets of the BioSecure Multimodal Database.

URL and following the instructions provided on the screen).
- **Data Set 2 (DS2)**, acquired in an office environment (**desktop**) using a standard PC and a number of commercial sensors under the guidance of a human acquisition supervisor.
- **Data Set 3 (DS3)**, acquired using **mobile** portable hardware under two acquisition conditions: indoor and outdoor. Indoor acquisitions were done in a quiet room, whereas outdoor acquisitions were recorded in noisy environments (office corridors, the street, etc.), allowing the donor to move and to change his/her position.

The BMDB has been designed to be representative of the population that would make possible use of biometric systems. As a result of the acquisition process, about 40 percent of the subjects are between 18 and 25 years old, 20-25 percent are between 25 and 35 years old, 20 percent of the subjects are between 35 and 50 years old, and the remaining 15-20 percent are above 50 years old, see Table 3. In addition, the gender distribution was designed to be as balanced as possible, with no more than 10 percent difference between male and female sets. Metadata associated with each subject were also collected to allow experiments regarding specific groups. The available information includes: age, gender, handedness, manual worker (yes/no), visual aids (glasses or contact lenses/none), and English proficiency. In Table 3 and Fig. 1, the actual statistics of the three data sets of the BioSecure Multimodal Database are shown. Other features are as follows:

- Two acquisition sessions in order to consider time variability (separated between one and three months).
- Multimodal data representative of realistic scenarios, with simple and quick tasks per modality (e.g., utterances of PINs or short sentences, still face images, handwritten signatures, hand images using a camera, etc.) instead of long tasks that are not appropriate in a real multimodal scenario (e.g., handwriting, hand images using a document scanner, etc.).
- Several biometric samples per session.
- Homogeneous number of samples and sessions per subject.
- Cross-European diversity (language, face, etc.) and site variability, achieved through the acquisition across different European institutions.
- Acquisition of the same trait with different biometric sensors in order to consider sensor variability.

The three data sets of BMDB include a common part of audio and video data, described in Table 4, which is comprised of still face images and talking face videos. Also, signature and fingerprint data have been acquired both in DS2 and DS3. Additionally, hand and iris data were acquired in DS2. All this information is summarized in Table 5, which is further detailed in the sections dedicated to each data set.

### 3.2 Data Set 1 (DS1)—Internet

The purpose of DS1 is to acquire material over the Internet without human supervision. For DS1, the acquisition protocol consisted of the common part described in Table 4. Therefore, the modalities included in DS1 are: voice and face. The first session was acquired using a PC provided by the acquisition center, with some guidance to let donors to become familiar with the acquisition and to correctly perform it, whereas the second session was allowed to take place over an uncontrolled scenario (at donor's home or office using appropriate hardware). In most of the cases, both sessions were acquired with the provided PC. Note that the speech information acquired enables both text-dependent (PINs) and text-independent speaker recognition experiments [27]. Digits are the same between sessions, enabling text-dependent speaker recognition based on digits. On the other hand, speakers also utter various sentences, with the sentences

TABLE 4
Common Contents of Audio and Video in the Three Data Sets

| Modality | Samples | # Samples |
|---|---|---|
| Face still | 2 still frontal face images | 2 image files |
| AV | 4 digits PIN code, 2 repetitions* from a set of 100 different codes, in English | 2 video files |
| AV | 4 digits PIN code, 2 repetitions* from a set of 10 different codes, in national language | 2 video files |
| AV | Digits from 0 to 9 in English | 1 video file |
| AV | 2 different phonetically rich sentences** in English | 2 video files |
| AV | 2 different phonetically rich sentences** in national language | 2 video files |
| | | 11 |

* The same PIN code between datasets and sessions.
** Different sentences between datasets and sessions.

TABLE 5
General Features of the Database and Modalities Acquired in Each Data Set

General features of the database

|  | DATA SET 1 (DS1) | DATA SET 2 (DS2) | DATA SET 3 (DS3) |
|---|---|---|---|
| Subjects | 971 | 667 | 713 |
| Sessions | 2 | 2 | 2 |
| Supervisor | No | Yes | Yes |
| Conditions | Over the **Internet** | Standard office (**desktop**) | **Mobile** indoor and outdoor |
| Hardware | PC, webcam, microphone | PC, commercial sensors | Portable devices |

Biometric data for each user and for each session in the database

| MODALITY | DATA SET 1 (DS1) | DATA SET 2 (DS2) | DATA SET 3 (DS3) | # SAMPLES |
|---|---|---|---|---|
| **Common AV - indoor** | 11 samples | 11 samples | 11 samples | 33 |
| - Audio-video | 4 PIN | 4 PIN | 4 PIN | |
| | 4 sentences | 4 sentences | 4 sentences | |
| | 1 digits sequence | 1 digits sequence | 1 digits sequence | |
| - Face still (webcam) | 2 frontal face images | 2 frontal face images | 2 frontal face images | |
| **Common AV - outdoor** | - | - | 11 samples | 11 |
| - Audio-video | | | 4 PIN | |
| | | | 4 sentences | |
| | | | 1 digits sequence | |
| - Face still (webcam) | | | 2 frontal face images | |
| **Signature** | - | 25 samples | 25 samples | 50 |
| | | 15 genuine | 15 genuine | |
| | | 10 imitations | 10 imitations | |
| **Fingerprint - thermal** | - | 12 samples ($3 \times 2 \times 2$) | 12 samples ($3 \times 2 \times 2$) | 24 |
| **Fingerprint - optical** | - | 12 samples ($3 \times 2 \times 2$) | - | 12 |
| **Iris** | - | 4 samples ($2 \times 2$) | - | 4 |
| **Hand - digital camera** | - | 8 samples ($2 \times 4$) | - | 8 |
| **Face still - digital camera** | - | 4 samples ($2 + 2$) | - | 4 |
| # SAMPLES | 11 | 76 | 59 | 146 |

being different in each session (see Table 4), thus also enabling text-independent speaker recognition.

The acquisition of DS1 was performed by connecting to an URL using an Internet browser and following the instructions provided on the screen. The acquisition was performed using a standard Webcam with microphone. In order to achieve realistic conditions, the use of no specific Webcam was imposed. The appearance of the graphical user interface of the application for the acquisition is shown in Fig. 2. Fig. 2a represents the user interface prepared for the acquisition of an audiovisual sample. The left panel shows the instructions to the donor, while the right panel shows the Webcam stream. Fig. 2b shows the graphical user interface just after an audiovisual sample has been acquired. The sample is presented to the donor in order to be validated before it is sent to the server. In Fig. 2c, a frontal still image has just been acquired. The donor has to adjust the position of his face to fit the overlaid "virtual glasses and chin" mask in order to normalize the pose.

### 3.3 Data Set 2 (DS2)—Desktop

The scenario considered for the acquisition of DS2 was a standard office environment. The acquisition was carried out using a desktop PC and a number of sensors connected to the PC via the USB or Bluetooth interface. Acquisition is managed by a supervisor, who is in charge of the following activities: 1) training of the contributor in case of difficulties in using a particular sensor; 2) validation of every acquired biometric sample, allowing a sample to be acquired again if it has not been properly acquired (e.g., wrong finger in the case of fingerprint acquisition or wrong utterance in the case of speech acquisition); and 3) guidance of the acquisition process by remembering the steps of the acquisition protocol and pointing out the sensor involved.

The modalities included in DS2 are: voice, face, signature, fingerprint, hand, and iris. Hardware used for the acquisition included a Windows-based PC with a USB hub, and the biometric sensors specified in Fig. 3. An example of the setup used by UPM is shown in Fig. 4, and the data acquired in DS2 is described in Table 6.

A specific application was developed for the acquisition of DS2, aiming to provide a common and homogeneous working interface for all the sites participating in the acquisition. This software tool allowed the recapture of samples until they exhibited satisfactory quality, the correction of invalid or missing samples, and the inclusion of new users or sessions at any point of the acquisition process. In

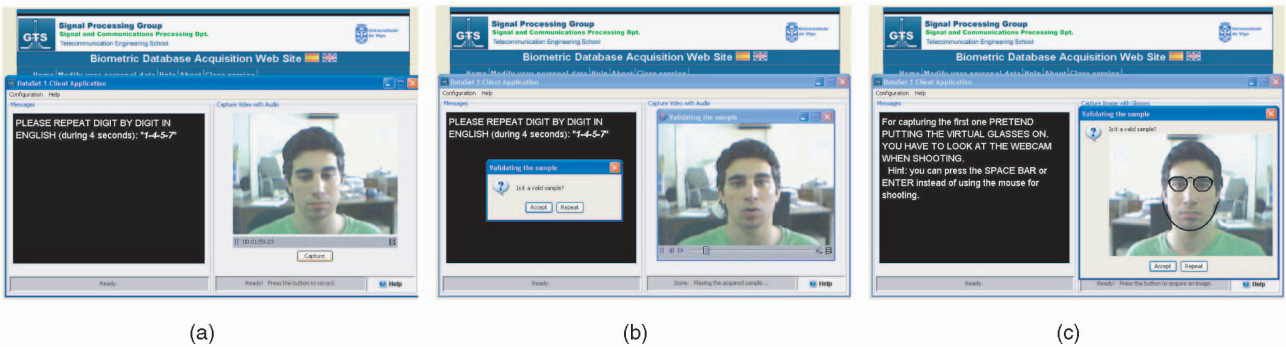

Fig. 2. Graphical user interface of the acquisition application for DS1.

Fig. 5, the screen captures of the main module and the different modules of the DS2 acquisition software are shown.

### 3.4 Data Set 3 (DS3)—Mobile

The aim of DS3 is to capture several modalities on mobile platforms. The modalities acquired in DS3 are: face, voice, fingerprint, and signature. For the common audio-video recordings, each session is comprised of two acquisition conditions, indoor and outdoor, performed during the same day. Hardware devices used for the acquisition include the biometric sensors specified in Fig. 6, and the data acquired in DS3 is described in Table 7. A supervisor was in charge of managing each acquisition session, providing appropriate training to the contributors in case of difficulties with a sensor, validating every acquired biometric sample, and allowing reacquisition in case of a sample that has not been properly acquired.

The two acquisition conditions considered for audio-video recordings are intended to be comprised of different sources of variability. "Indoor" acquisitions were done in a quiet room, just changing the position between each audio-video sequence. "Outdoor" acquisitions were recorded in noisy environments such as building corridors, the street, etc., allowing the donor to move and to change position during and between each audio-video sequence. For signature and fingerprint, only indoor data were acquired, which can be considered as degraded with respect to DS2, as signatures and fingerprints were acquired while standing and holding the PDA.

The acquisition with Samsung Q1 was carried out using the same application used for DS2, adapted and limited to the audio-video acquisition task. For the acquisition of data with PDA, a specific software tool was developed. In Fig. 7, screen captures of the main module and the different modules of the DS3 acquisition software for PDA are shown.

### 3.5 Acquisition Guidelines and Validation of Acquired Data

A data validation step was carried out to detect invalid samples within the three data sets. We should distinguish between valid low-quality samples and invalid samples. Low-quality samples are acceptable as long as the acquisition protocol is strictly followed and the different biometric sensors are correctly used (e.g., blurred iris images, dry or wet fingerprint images, etc.). These samples were not removed from the database since they represent real-world samples that can be found in the normal use of a biometric system. On the other hand, invalid samples are those that do not comply with the specifications given for the database

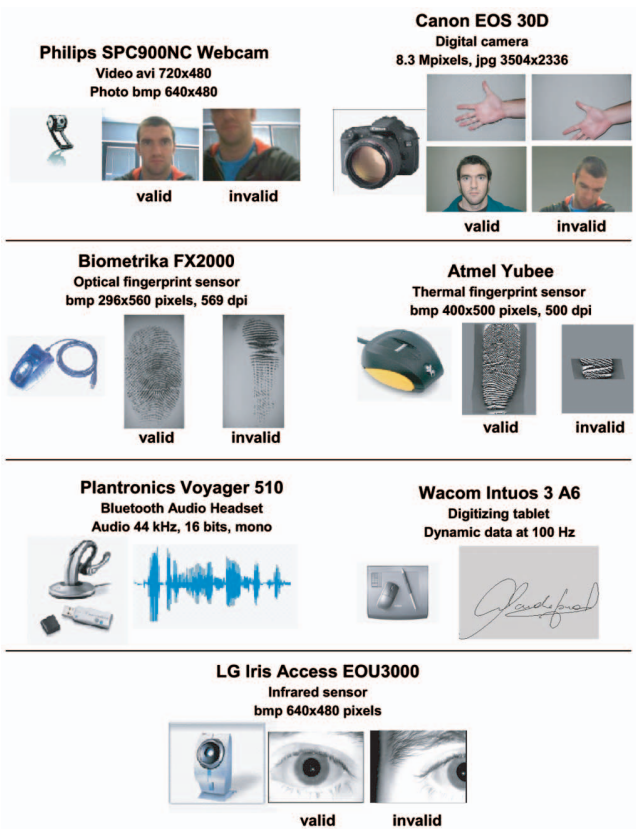

Fig. 3. Hardware devices used in the acquisition of DS2 together with acquisition samples.

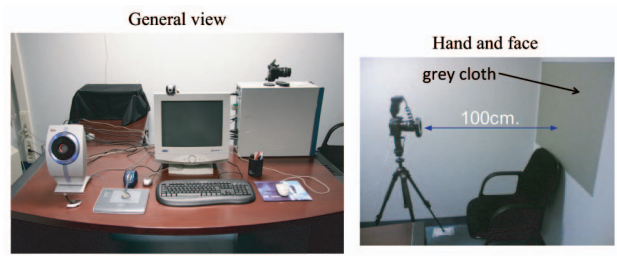

Fig. 4. Example of the setup used by UPM for the acquisition of DS2.

TABLE 6
Acquisition Protocol of DS2

| Modality | Sensor | Samples | # Samples |
|---|---|---|---|
| Signature | Tablet | 5 genuine signatures of donor $n$ | 5 text files |
| | | 5 dynamic imitations of donor $n$-1 ($n$-3 in session 2) | 5 text files |
| | | 5 genuine signatures of donor $n$ | 5 text files |
| | | 5 dynamic imitations of donor $n$-2 ($n$-4 in session 2) | 5 text files |
| | | 5 genuine signatures of donor $n$ | 5 text files |
| Face Still | Webcam | Common still frontal face images (see Table IV) | 2 image files |
| AV | Webcam+headset | Common audio-video part (see Table IV) | 9 video files |
| Iris image | Iris camera | (Right eye, Left eye) $\times$ 2 | 4 image files |
| Fingerprint | Optical | ( (Thumb $\rightarrow$ index $\rightarrow$ middle) $\times$ 2 hands {right $\rightarrow$ left} ) $\times$ 2 | 12 image files |
| | Thermal | ( (Thumb $\rightarrow$ index $\rightarrow$ middle) $\times$ 2 hands {right $\rightarrow$ left} ) $\times$ 2 | 12 image files |
| Hand | Digital camera | (Right hand $\rightarrow$ Left hand) $\times$ 2 without flash | 4 image files |
| | | (Right hand $\rightarrow$ Left hand) $\times$ 2 with flash | 4 image files |
| Face Still | Digital camera | 2 photos without flash $\rightarrow$ 2 photos with flash | 4 image files |
| | | | **76 files** |

(e.g., wrong PIN, forgery of a wrong signature, donor's head or hand out of frame, etc.).

The first stage of the validation process was carried out during the acquisition itself. A human supervisor was in charge of validating every acquired biometric sample, being recaptured if it did not meet the specified quality standards. After completion of the acquisition, a second validation step of the three data sets was carried out again manually by a human expert. The following validation criteria and acquisition guidelines were given to the supervisors in charge of validation of the data:

- Face acquisition (both face still and video): Donor pose should be frontal (looking straight into the camera) and with neutral expression, and donor's head should not be out of frame. In video files, audio and video fields should be synchronized. Blurred images are not discarded, unless the face is clearly nonvisible.
- Iris acquisition: Glasses (if any) should be removed before acquisition, but contact lenses are acceptable. A part of donor's eye falling out of frame or eye closed are not allowed.
- Hand acquisition: The hand pose is with wide open fingers. Fingers too close together or part of the hand out of frame are not allowed.
- Fingerprint acquisition with the optical sensor: The contact surface of the device should be cleaned after each donor session. For fingerprint acquisition with the thermal sensor, as it is difficult to use correctly, the donor was allowed to try multiple times before the first acquisition. Very low-quality fingerprints or very small-size images due to improper use of the sensor are not allowed.
- Signature acquisition: Donors were asked to sign naturally (i.e., without breaks or slowdowns). For

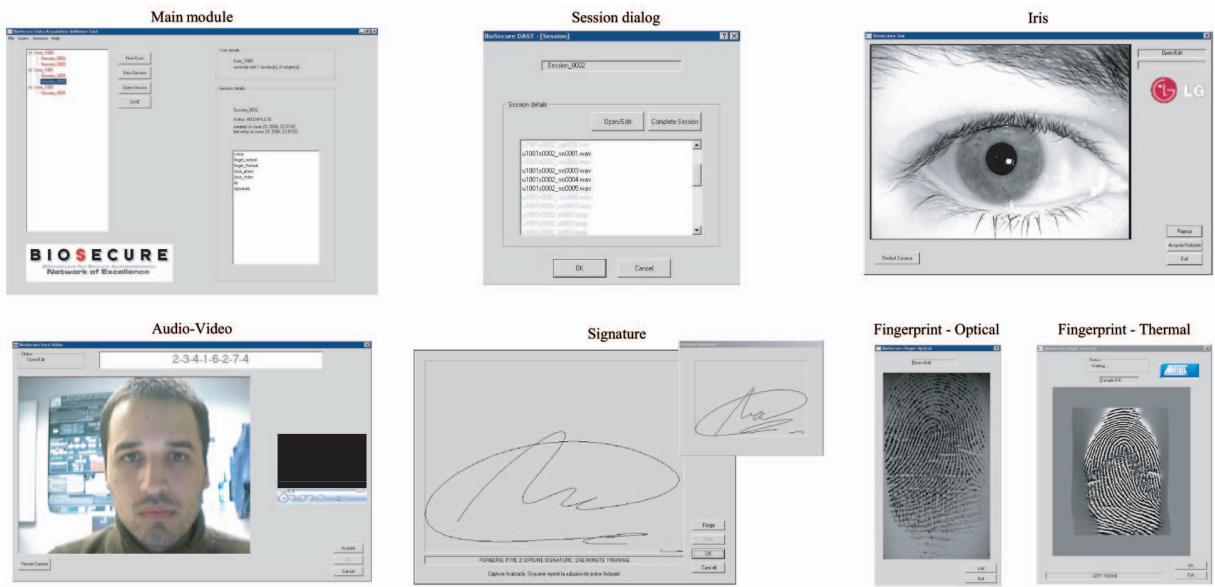

Fig. 5. Screen captures of the main module (top left) and the different modules of the DS2 acquisition software.

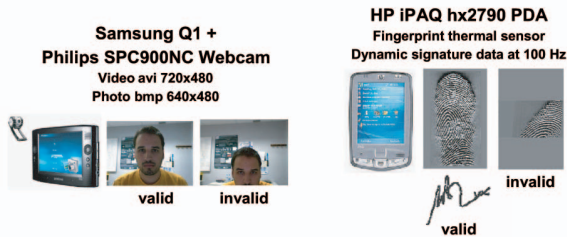

Fig. 6. Hardware devices used in the acquisition of DS3 together with acquisition samples.

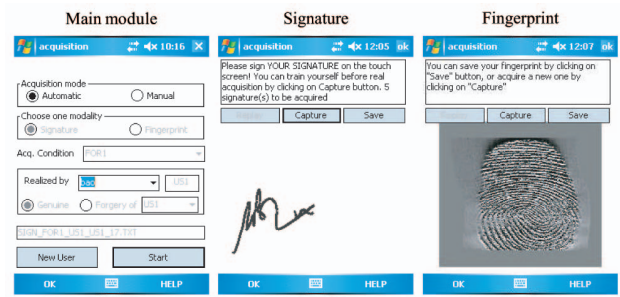

Fig. 7. Screen captures of the main module (left) and the different modules of the DS3 acquisition software for PDA.

impostor realizations, signature to be imitated could be replayed on the screen with the dynamic process and the donor was allowed to train before forging.

### 3.6 Problems Encountered During the Acquisition

A list of problems encountered has been gathered during the acquisition of the database due to the feedback provided by the donors and/or the different acquisition sites. Concerning usability, the most relevant are:

- The Yubee thermal fingerprint sensor was difficult to use, requiring many trials to get a reasonably good fingerprint capture. This sensor also caused annoyance in some users due to Failure to Enroll error.
- The autofocusing function of the iris camera was not always working, needing several trials to get a reasonably sharp iris image capture. Focus on iris scanner sometimes did not always correspond to the acquisition instant.
- Lighting had influence on the quality of acquired iris images in some cases, requiring reducing the overall illumination of the room.
- For some volunteers doing the face pictures, there were too many reflections on their glasses.
- During swiping the finger on the fingerprint sensor of the PDA, there was a risk of touching the screen or the button just below the sensor. This could end the capture mode of the software, requiring a new acquisition.

## 4 COMPATIBILITY OF THE BIOSECURE MULTIMODAL DATABASE WITH OTHER DATABASES

Several BioSecure partners have considerable experience in biometric database collection (e.g., MCYT [20], XM2VTS [23], BIOMET [16], MyIDEA [15], MBioID [17], BioSec [14], or BT-DAVID [25]). In recent years, various multimodal databases have been collected, mainly in academic environments. As a result, permanent staff members and some students have participated in the biometric data acquisition for these databases over several years. The existence of these data enables the number of sessions for a subset of such donors to be increased, therefore allowing studies of long-term variability. Specifically, the BioSecure MDB shares individuals with:

- The **BioSec** database, which is comprised of 250 subjects. The two databases have in common the following traits: optical/thermal fingerprint, face, speech, and iris. Both databases have in common 25 subjects, separated by about two years.
- The **BiosecurID** database, which is comprised of 400 subjects. The two databases have in common the following traits: optical/thermal fingerprint, face, hand, signature, speech, and iris. Both databases have in common 31 subjects, separated by about one year.

It should also be noted that the devices and protocol of some of the traits acquired in the BioSecure MDB are compatible with other existing databases, so it can be combined with portions of them to increase the number of available subjects, specifically: the above-mentioned BioSec and BiosecurID, the MCYT database (signature), and the MyIDEA database (signature and fingerprint).

TABLE 7
Acquisition Protocol of DS3

| Modality | Sensor | Condition | Samples | # Samples |
|---|---|---|---|---|
| Signature | PDA | Indoor | 5 genuine signatures of donor $n$ | 5 |
| | | | 5 dynamic forgeries of donor $n$-1 ($n$-3 in session 2) | 5 |
| | | | 5 genuine signatures of donor $n$ | 5 |
| | | | 5 dynamic forgeries of donor $n$-2 ($n$-4 in session 2) | 5 |
| | | | 5 genuine signatures of donor $n$ | 5 |
| Fingerprint | PDA (thermal) | Indoor | ( (Thumb → index → middle) × 2 hands {right → left} ) × 2 | 12 |
| Face Still | Q1+ | Indoor | Common still frontal face images (see Table IV) | 2 |
| AV | webcam | Indoor | Common audio-video part (see Table IV) | 9 |
| Face Still | Q1+ | Outdoor | Common still frontal face images (see Table IV) | 2 |
| AV | webcam | Outdoor | Common audio-video part (see Table IV) | 9 |
| | | | | 59 |

## 5 BIOSECURE MULTIMODAL EVALUATION CAMPAIGN

The BMEC [10] was conducted during 2007 as a continuation of the BioSecure Multimodal Database acquisition. At the time the evaluation was done, the database was not fully completed. Therefore, the BMEC was done on a selected subset of the database in order to show its consistency and value. Two different scenarios were defined for the Evaluation: an Access Control scenario on DS2 data and a degraded Mobile scenario on data from DS3. For these scenarios, different multimodal experiments were carried out using matching scores generated by several reference monomodal systems [28], in order to compare different fusion schemes both by BioSecure partners and by external researchers who responded to a public call for participation. In addition, monomodal evaluations on four modalities from DS3 were also performed. Since it is not within the purposes of this paper to describe in detail the BioSecure Multimodal Evaluation Campaign or its results, we will give only a brief outline of the experimental protocols used in these evaluations, together with some selected results, so the reader can have a reference experimental framework as well as baseline results.

### 5.1 Access Control Scenario Evaluation

The aim of the *DS2 Evaluation (Access Control scenario evaluation)* was to compare multimodal biometric fusion algorithms assuming that the environment is well controlled and the users are supervised, using data from BMDB DS2 (desktop data set). Three biometric traits were employed: still face images, fingerprint, and iris. Together with the scores computed using the reference systems, a set of quality measures was also extracted from the biometric samples. An LDA-based face verifier using Fisher Linear Discriminant projection [29] was used as the face reference system, with correlation as similarity measure (between two projected features). Different face quality measures were computed using the Omniperception proprietary Affinity SDK,[1] including measures like contrast, brightness, etc. For the fingerprint modality, the NIST minutiae-based fingerprint system was used [30], whereas the quality measure was based on averaging local gradients [31]. Minutiae are detected using binarization and thinning operations, and for each minutia, its position, direction, type, and quality are computed. Matching is done by looking for correspondences between two pairs of minutia, one pair from the template fingerprint and one pair from the test fingerprint. The fingerprint matcher is rotation and translation-invariant. As the iris reference system, a variant of the Libor Masek's system was used [32] and three different iris quality measures were computed: texture richness, difference between iris and pupil diameters, and proportion of iris used for matching. Masek's system employs the circular Hough transform for iris boundaries' detection, and the Daugman's rubber sheet [33] model for normalization of the iris region. Feature encoding is implemented by using 1D Log-Gabor wavelets and phase quantization to four levels using the Daugman method [33]. For matching, the Hamming distance is chosen as a metric for recognition.

1. http://www.omniperception.com.

For this scenario, data from 333 individuals of DS2 has been used. Two sets of scores are computed, one for development and another for testing. There are 51 individuals in the development set (provided to the evaluation participants to tune their algorithms), 156 individuals in the test set (sequestered and only used by the evaluation organizers on the already tuned algorithms), and 126 individuals set as an external population of zero-effort impostors. For each person, four samples per modality are available (two per session, see Table 6). The first sample of the first session is used as a template. The remaining samples are considered as query data (the other from Session 1 and the two from Session 2). The evaluated fusion algorithms were tested using only the query samples of Session 2. This testing was done using as impostors the external population of 126 individuals set as zero-effort impostors. This external population was not used elsewhere. In this way, a fusion algorithm will not have already "seen" the impostors during its training stage (for which data from Session 1 can be used), thus avoiding optimistic performance bias.

Two types of evaluations were proposed in this scenario:

- *Quality-based evaluation*, aimed at testing the capability of fusion algorithms to cope with template and query biometric signals acquired with different devices, by exploiting quality information in the information fusion process [34]. Face and fingerprint modalities are considered in this case. Face images collected with the Webcam (referred to as low-quality data) and the digital camera (high-quality data) are denoted as *fa1* and *fnf1* streams, respectively. Fingerprint data include images from the two sensors used in DS2. They are denoted as *fo* (optical) and *ft* (thermal). The case where templates and query images are acquired with different devices is also considered. For the face modality, this is denoted as *xfa1* stream, i.e., the templates acquired with *fnf1* (digital camera) and the queries with *fa1* (Webcam). Similarly, for the fingerprint modality, the cross-device stream is denoted as *xft*, i.e., the templates acquired with *fo* (optical) and the queries with *ft* (thermal). It should be noted that the purpose of these cross-device experiments was just to evaluate the effect of matching biometric signals coming from different devices, without any special adjustment of the preprocessing or matching steps to deal with this issue. Therefore, in this case, the order of the samples (i.e., template and query) has no relevant impact.
- *Cost-based evaluation*, aimed at achieving the best performance with a minimal cost of acquiring and processing biometric information. The use of each biometric trait is associated with a given cost. Only face images from the Webcam (*fa1*) and fingerprint images from the thermal sensor (*ft*) are used, together with iris images (denoted as *ir1*). No cross-device experiments are conducted in this case. In the evaluation, for each modality used in the fusion, one cost unit is charged.

### 5.2 Mobile Scenario Evaluation

The *DS3 Evaluation (Mobile scenario evaluation)* was aimed at comparing biometric recognition algorithms assuming that the data are acquired using mobile devices and the users are not supervised, using data from BMDB DS3 (mobile data set). For multimodal experiments, 2D face video, fingerprint, and signature data were used. Monomodal experiments on 2D face video, fingerprint, signature, and talking face data were also carried out. 2D face video scores for the multimodal evaluation were generated using an eigenface-based approach developed by Bogazici University [35]. It uses the standard eigenface approach to represent faces in a lower dimensional subspace. All the images used by the system are first normalized. The face space is built using a separate training set and the dimensionality of the reduced space is selected such that 99 percent of the variance is explained by the Principal Component Analysis. After that, all the target and test images are projected onto the face space. Then, the L1 norm is used to measure the distance between the projected vectors of the test and target images. For the fingerprint modality, the NIST fingerprint system was used [30]. The signature reference system was developed by GET-INT (currently TELECOM & Management SudParis) and is based on Hidden Markov Models [36], [37], [38]. Twenty-five dynamic features are extracted at each point of the signature. Signatures are modeled by a continuous left-to-right HMM [39], by using, in each state, a continuous multivariate Gaussian mixture density. The number of states in the HMM modeling the signatures of a given person is determined individually according to the total number of all the sampled points available when summing all the genuine signatures that are used to train the corresponding HMM. Matching is done using the Viterbi algorithm [39].

For this scenario, 480 individuals from DS3 were considered. A set of 50 individuals was used for development, whereas the remaining 430 were used for testing. Two different experiments were carried out in the multimodal evaluation of this scenario, one using random signature forgeries and the other using skilled signature forgeries. For the 2D face modality, two indoor video samples of the first session were used as templates, whereas two outdoor video samples of the second session were used as query data. For the fingerprint modality, two samples of the first session were used as template data and two samples of the second session as query data. Signature templates were generated using five genuine signatures of the first session. As query data, genuine signatures of the second session and forgeries acquired during both sessions are used.

### 5.3 Baseline Results

We plot in Figs. 8 and 9 the verification results using the scores generated for the DS2 and DS3 multimodal evaluations. For the DS2 evaluation, results are shown using only query samples of the Session 2, as testing of the algorithms is done only using query samples of this session.

By looking at Fig. 8, it is observed that the performance of the face modality is degraded when using the Webcam, both if we match two images from the Webcam and if we mismatch one Webcam and one digital camera image. This is not true for the fingerprint modality, where a significant degradation is observed in the *xft* stream. As revealed in previous studies [40], [41], matching fingerprint images coming from different sensors has severe impact on the recognition rates due to variations introduced by the sensors (e.g., see fingerprint images of Figs. 3 or 5).

Regarding the results on DS3 data shown in Fig. 9, it is remarkable the degradation of performance of the face modality with respect to the *fa1* stream of DS2, in which the same Webcam is used for the acquisition. The more challenging environment of DS3, including outdoor acquisitions in noisy ambience, has a clear impact on the performance of the face modality (e.g., see Fig. 10). The face reference system is based on the PCA approach, which is not adequate to cope with the huge illumination variability. It is worth noting that this is not observed in the fingerprint modality, where no degradation is observed with respect to the *ft* fingerprint stream of DS2. Since the two sensors are based on the same (thermal) technology, the quality of fingerprint images is not affected by the differences in the acquisition conditions between DS2 and DS3.

The demographic statistics of the development and test sets used in the Mobile scenario evaluation are quite similar in terms of gender, age, and handedness of the donors. The main difference is found in the visual aids. As there are more people wearing glasses in the test database, we can suppose that the 2D face test database is "more difficult" than the corresponding development database. This is mirrored in the results of Fig. 9. For the other modalities (fingerprint and signature), performance results are observed to be very close on both sets.

## 6 RESEARCH AND DEVELOPMENT USING THE BIOSECURE MULTIMODAL DATABASE

Some of the research and development directions that can be followed using the database have been already put forward through this paper. In this section, we summarize these directions. It has to be emphasized that the BioSecure Multimodal Database includes new challenging scenarios not considered in existing biometric databases. This new database is unique in that it includes hundred of users acquired simultaneously under different environmental conditions and using multiple sensors for the same modality. It allows uses such as:

- Novel research in the available modalities or in multibiometric systems, with very tight confidence intervals on results thanks to the size of the database.
- Evaluation of multibiometric algorithms using a common framework and real multimodal data. We should remark that the high error rates of the baseline results presented in Section 5.3 (between 5 and 10 percent EER for the best cases) leave a lot of room for improvement, either with individual modalities or with multibiometric approaches.
- Evaluation of sensor interoperability on a large scale in traits acquired with several devices (face, fingerprint, speech, and signature) [42], [43] due to the amount of available data, as was done in the BMEC [10].

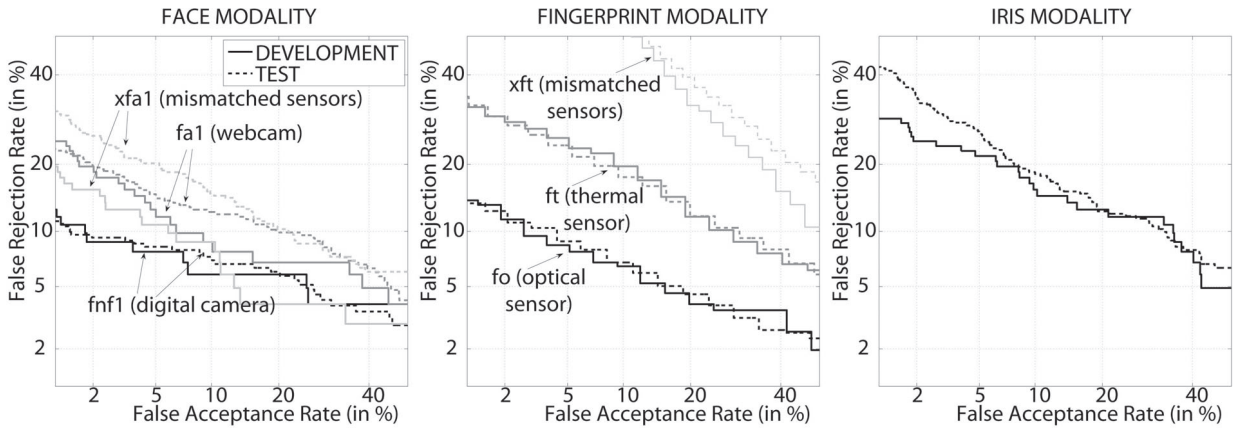

Fig. 8. **Access control scenario evaluation (BMDB DS2, desktop data set).** Baseline results of the face, fingerprint, and iris modalities.

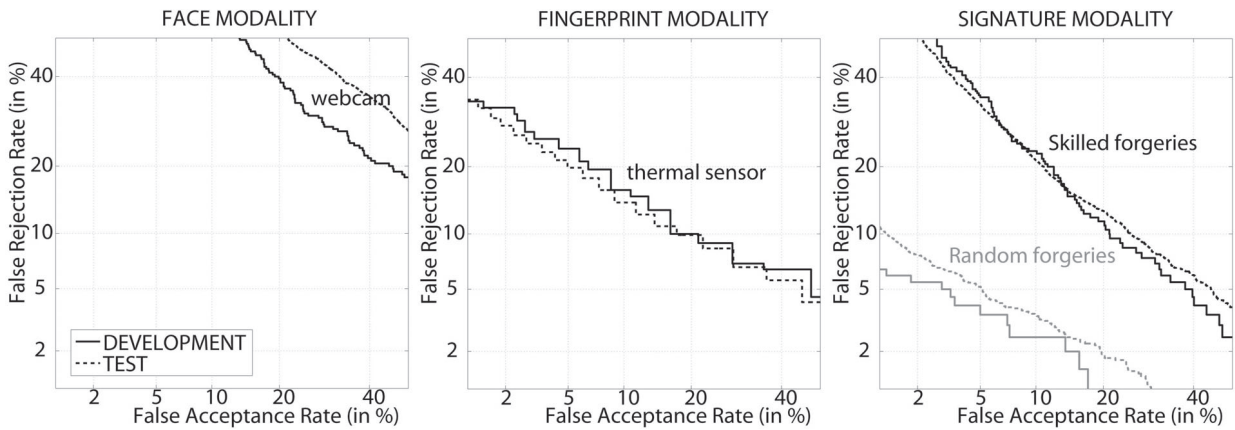

Fig. 9. **Mobile scenario evaluation (BMDB DS3, mobile data set).** Baseline results of the face, fingerprint, and signature modalities.

- Study of differences in system performance due to environmental variations: over the Internet, in an office environment with a desktop PC, or with mobile platforms in an indoor/outdoor environment.
- Evaluation of several multibiometric realistic scenarios, e.g., bimodal passport using face and fingerprint, Internet-based access using face and voice, mobile-based access with face and fingerprint, etc.
- Study of time variability in biometric systems and template update techniques [44]. Research can be done on the short term (considering data from the same session), on the medium term (considering data from the different sessions), or on the long term (considering common data from other databases, as mentioned in Section 4).
- Evaluation of potential attacks to monomodal or multimodal systems [45].
- Effect of the different acquisition scenarios/devices on the quality of acquired samples and its impact on the recognition performance [46], [47].
- Biometric studies depending on demographic information such as age [48], gender [49], handedness, state of the hand (manual workers), or visual aids.
- Cross-European diversity and site variability studies in terms of language (speech), appearance (face), etc.

## 7 LEGAL ISSUES AND DISTRIBUTION

Directive 95/46/EC of the European Parliament and the Council of 24 October 1995 sets the European requirements on the protection of individuals with regard to the processing of personal data and on the free movement of such data. According to this directive, biometric data are considered as "personal data." Based on this regulation, donors were asked to read and sign a consent agreement before starting the acquisition process, which included comprehensive information about the motivation and planned use of the biometric data, with pointers to the security measures applied to protect these data.

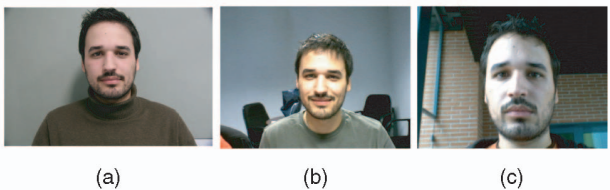

Fig. 10. Variability between face samples acquired with different sensors and in different environments. Face images are plotted for (a) indoor digital camera (from DS2, left, resized at 20 percent), (b) indoor Webcam (from DS2, medium), and (c) outdoor Webcam (from DS3, right).

Other personal data were acquired and stored securely and independently of the biometric data, including: name, contact details, age, gender, handedness, manual worker, vision aids, and English proficiency. These nonbiometric data are managed by the BioSecure partner involved in the acquisition of the donor at hand (also referred to as the "controller" in the Directive 95/46/EC). In any subsequent use or transfer of data, only the raw biometric data plus the fields {age, gender, handedness, visual aids, manual worker, and English proficiency} are considered, without any link to identities of the donors (i.e., name and contact details).

The BioSecure Multimodal Database will be distributed during 2008. The distribution will be managed by the recently created BioSecure Association.[2] This Association is concerned with the use and dissemination of the results generated within the BioSecure Network of Excellence involving Intellectual Property Rights.

## 8 CONCLUSION

The existence of evaluation procedures and databases is crucial for the development of biometric recognition systems. It is often the existence of data sets with new challenging conditions which drives research forward. Public biometric databases allow the creation of common and repeatable benchmarks and algorithms, so that new developments can be compared with existing ones. However, biometric database collection is a time and resource-consuming process, specially in the case of multimodal databases. As a result, most of the existing biometric databases typically include only one or two modalities. Fortunately, in recent years, important efforts have been directed to collecting real multimodal databases, mainly in the framework of collaborative national or international projects, resulting in a number of databases available or in the process of completion.

In this contribution, the recently acquired BioSecure Multimodal Database is presented, together with a brief description of previous work in the domain of multimodal biometric database acquisition. This database is the result of an important collaborative effort of 11 European partners of the BioSecure NoE. It includes new challenging acquisition conditions and features not present in existing databases. It is comprised of three different data sets with more than 600 common individuals captured in two sessions: 1) one data set acquired over the Internet, 2) another one acquired in an office environment with a desktop PC, and 3) the last one acquired with mobile devices in indoor/outdoor environments. The three data sets include a common part of audio and video data which comprise still images of frontal face and talking face videos acquired with a Webcam. Additionally, the second data set includes still face (with a digital camera), signature, fingerprint (with two different sensors), and hand and iris data, and the third one also includes signature and fingerprint data. Also worth noting, the BioSecure Multimodal Database shares a number of individuals with other multimodal databases acquired across several years, allowing studies of long-term variability.

The new challenging acquisition conditions of this database will allow the evaluation of realistic multimodal scenarios, as was done in the recently conducted BMEC. A brief description of this evaluation together with baseline results of individual modalities from the database was also provided in this paper, from which a number of experimental findings related to biometric recognition using mismatched devices and heterogeneous acquisition conditions have been highlighted.

## ACKNOWLEDGMENTS

This work has been supported by the European Network of Excellence (NoE) BioSecure—Biometrics for Secure Authentication—and by the National Projects of the Spanish Ministry of Science and Technology (TEC2006-13141-C03-03, TEC2006-03617/TCM, TIC2002-04495-C02, and TEC2005-07212) and the Italian Ministry of Research. The postdoctoral research of author J. Fierrez is supported by a Marie Curie Outgoing International Fellowship. The authors F. Alonso-Fernandez and M.R. Freire are supported by FPI Fellowships from Comunidad de Madrid. The author J. Galbally is supported by an FPU Fellowship from Spanish MEC. Authors Josef Kittler and Norman Poh are supported by the Advanced Researcher Fellowship PA0022_121477 of the Swiss National Science Foundation and by the EU-funded Mobio project grant IST-214324. The author J. Richiardi is supported by the Swiss National Science Foundation. The authors also thank the support and assistance of (in alphabetical order): Dr. Manuele Bicego (UNISS), Prof. Enrico Grosso (UNISS), Dr. Andrea Lagorio (UNISS), Aurélien Mayoue (TELECOM & Management SudParis), Dijana Petrovska (TELECOM & Management SudParis), Ms. Ajita Rattani (UNISS), Florian Verdet (UNIFRI). The authors would also like to thank the anonymous donors who have contributed to this database.

---

2. More information to be found at: http://biosecure.it-sudparis.eu/AB.

**Javier Ortega-Garcia** received the PhD degree "cum laude" in electrical engineering in 1996 from the Universidad Politecnica de Madrid, Spain. He is the founder and co-director of the Biometric Recognition Group—ATVS. He is currently a full professor at the Escuela Politecnica Superior, Universidad Autonoma de Madrid. He is a senior member of the IEEE. More details about his research can be found at http://atvs.ii.uam.es.

**Julian Fierrez** received the PhD degree in telecommunications engineering from the Universidad Politecnica de Madrid, Spain, in 2006. Since 2002, he is affiliated with the Universidad Autonoma de Madrid in Spain, where he currently holds a Marie Curie postdoctoral fellowship, part of which has been spent as a visiting researcher at Michigan State University. He is a member of the IEEE. More details about his research can be found at http://atvs.ii.uam.es.

**Fernando Alonso-Fernandez** received the PhD degree (cum laude) in electrical engineering from the Universidad Politecnica de Madrid, Spain, in 2008. Since 2004, he has been with the Universidad Autonoma de Madrid in Spain, where he currently holds a Juan de la Cierva postdoctoral fellowship of the Spanish MCINN. He is a member of the IEEE. More details about his research can be found at http://atvs.ii.uam.es.

**Javier Galbally** received the MSc degree in electrical engineering from the Universidad de Cantabria, Spain, in 2005. He is currently working toward the PhD degree at the Universidad Autonoma de Madrid, where he is an assistant researcher with the Biometrics Recognition Group—ATVS. He is a student member of the IEEE. More details about his research can be found at http://atvs.ii.uam.es.

**Manuel R. Freire** received the MSc degree in computer science from the Universidad Autonoma de Madrid in 2008. He is currently working toward the PhD degree at the Universidad Autonoma de Madrid, where he has been a research assistant with the the Biometric Recognition Group—ATVS since 2006. More details about his research can be found at http://atvs.ii.uam.es.

**Joaquin Gonzalez-Rodriguez** received the PhD degree "cum laude" in electrical engineering from the Universidad Politecnica de Madrid, Spain, in 1999. He is the founder and co-director of the Biometric Recognition Group—ATVS. Since May 2006, he has been an associate professor in the Computer Science Department at the Universidad Autonoma de Madrid, Spain. He is a member of the IEEE. More details about his research can be found at http://atvs.ii.uam.es.

**Carmen Garcia-Mateo** received the PhD degree (Hons) in telecommunications engineering from the Universidad Politécnica de Madrid, Spain, in 1993. She is a full professor in the field of discrete signal processing at the Universidad de Vigo. She is a member of the IEEE. More details about her research can be found at http://www.gts.tsc.uvigo.es.

**Jose-Luis Alba-Castro** received the PhD degree (Hons) in telecommunications engineering from the Universidad de Vigo in 1997. He is an associate professor of discrete signal processing, pattern recognition, image processing, and biometrics at the Universidad de Vigo. He is a senior member of the IEEE. More details about his research can be found at http://www.gts.tsc.uvigo.es.

**Elisardo Gonzalez-Agulla** received the telecommunications engineer degree from the Universidad de Vigo, Spain, in 2002, where he is currently working toward the PhD degree in the field of biometrics for Web applications. More details about his research can be found at http://www.gts.tsc.uvigo.es.

**Enrique Otero-Muras** received the telecommunications engineer degree from the Universidad de Vigo, Spain, in 2005. He is currently with a company related to biometrics and video-based security.

**Sonia Garcia-Salicetti** received the PhD degree in computer science from the University of Paris 6 (Université Pierre et Marie Curie, France) in December 1996. Since 2001, she has been an associate professor with the Interaction for Multimedia research team at the TELECOM & Management SudParis (Institut TELECOM). She is a member of the IEEE. More details about her research can be found at http://biometrics.it-sudparis.eu.

**Lorene Allano** received the PhD degree in computer science from the TELECOM & Management SudParis (Institut TELECOM) in January 2009. She is currently a postdoctoral researcher at Commissariat à l'Energie Atomique.

**Bao Ly-Van** received the PhD degree in system optimization and security from the TELECOM & Management SudParis (Institut TELECOM), Evry, France, in 2005. He is currently a freelance consultant in software engineering.

**Bernadette Dorizzi** received the PhD (thèse d'état) degree in theoretical physics from the University of Orsay (Paris XI—France) in 1983. She has been a professor at the TELECOM & Management SudParis (Institut TELECOM) since September 1989, as the head of the Electronics and Physics Department since 1995, where she is in charge of the Interaction for Multimedia research team. She is the coordinator of the BioSecure Association (http://biosecure.info). More details about her research can be found at http://biometrics.it-sudparis.eu.

**Josef Kittler** received the BA, PhD, and DSc degrees from the University of Cambridge, United Kingdom, in 1971, 1974, and 1991, respectively. He heads the Centre for Vision, Speech, and Signal Processing, Faculty of Engineering and Physical Sciences, University of Surrey, Guildford, United Kingdom. More details about his research can be found at http://www.ee.surrey.ac.uk/CVSSP.

**Thirimachos Bourlai** received the MSc degree (with distinction) in medical imaging, the PhD degree, and the postdoc in multimodal biometrics from the University of Surrey (UniS), United Kingdom, in 2002, August 2006, and August 2007, respectively. Since September 2007, he has been a postdoctoral researcher at the University of Houston, where he is engaged in research in the fields of cardiac pulse and stress detection.

**Norman Poh** received the PhD degree in computer science from the Swiss Federal Institute of Technology in Lausanne (EPFL), Switzerland, in 2006. Since then, he has been a research fellow at CVSSP, University of Surrey, United Kingdom. He is a member of the IEEE. More details about his research can be found at http://www.ee.surrey.ac.uk/CVSSP.

**Farzin Deravi** received the PhD degree from the University of Wales, Swansea, United Kingdom, in 1987. In 1998, he joined the Department of Electronics at the University of Kent, United Kingdom, where he is currently a reader in information engineering. He is a member of the IEEE. More details about his research can be found at http://www.ee.kent.ac.uk.

**Ming W.R. Ng** received the BSc (Hons) degree in electronic and electrical engineering from the Loughborough University of Technology, United Kingdom, and the PhD degree from the University of Kent, United Kingdom, in 2004. He is currently a research associate in the Department of Electronics at the University of Kent. More details about his research can be found at http://www.ee.kent.ac.uk.

**Michael Fairhurst** received the PhD degree from the University of Kent, Canterbury, United Kingdom, where he has been the head of the Department of Electronics until 2008. He is a member of the Biometrics Working Group and the UK Government's Biometrics Assurance Group. More details about his research can be found at http://www.ee.kent.ac.uk.

**Jean Hennebert** received the PhD degree in computer science from the Swiss Federal Institute of Technology, Lausanne, Switzerland, in 1998. Since 2007, he has been a professor with the Institute of Business Information Systems, University of Applied Sciences Western Switzerland, HES-SO//Wallis, Sierre, Switzerland. He is also affiliated as a lecturer with the University of Fribourg. He is a member of the IEEE. More details about his research can be found at http://iig.hevs.ch and at http://diuf.unifr.ch/diva.

**Andreas Humm** received the PhD degree in computer science from the University of Fribourg, Switzerland, in 2008. Since 2009, he has been a postdoctoral researcher in the multimedia engineering DIVA group of the Computer Science Department, University of Fribourg. He is a member of the IEEE. More details about his research can be found at http://diuf.unifr.ch/diva.

**Massimo Tistarelli** received the PhD degree in computer science and robotics from the University of Genoa, Italy, in 1991. He is currently a full professor in computer science and director of the Computer Vision Laboratory at the University of Sassari, Italy. He is a senior member of the IEEE. More details about his research can be found at http://biometrics.uniss.it/lecturers/tistarelli/bio_tistarelli.php and at http://www.architettura.uniss.it/article/articleview/455/1/62/.

**Linda Brodo** received the PhD degree in computer science from the University of Verona in 2003. She has been a researcher on the Faculty of Foreign Languages and Literatures at the University of Sassari, Italy, since 2004. She also collaborates with the Computer Vision Laboratory at the University of Sassari in the field of biometric face recognition.

**Jonas Richiardi** received the PhD degree in 2008 from the Signal Processing Institute, Swiss Federal Institute of Technology, Lausanne, Switzerland, where he is currently a postdoctoral researcher. He is also the cofounder of the research and development company PatternLab. He is a member of the IEEE. More details about his research can be found at http://scgwww.epfl.ch.

**Andrzej Drygajlo** received the PhD (summa cum laude) degree in electronics engineering from the Silesian Technical University, Gliwice, Poland, in 1983. He is the head of the Speech Processing and Biometrics Group at the Swiss Federal Institute of Technology at Lausanne. He is a member of the IEEE. More details about his research can be found at http://scgwww.epfl.ch.

**Harald Ganster** received the PhD degree (with distinction) in technical mathematics from the University of Technology Graz, Austria, in 1999. He is currently with the Institute of Digital Image Processing at Joanneum Research in Graz and is responsible for the field of biometrics. More details about his research can be found at http://www.joanneum.at.

**Federico M. Sukno** received the PhD degree in biomedical engineering from the University of Zaragoza in 2008. He is currently engaged in research on statistical shape models at Pompeu Fabra University, where he is an assistant professor in coding theory and analog electronics. More details about his research can be found at http://www.cilab.upf.edu.

**Sri-Kaushik Pavani** received the MS degree in electrical engineering from Texas Tech University in 2003. He is currently working toward the PhD degree in computer science and visual communication at the Universitat Pompeu Fabra, Barcelona. More details about his research can be found at http://www.cilab.upf.edu.

**Alejandro Frangi** received the PhD degree in biomedical image analysis from the Image Sciences Institute of the University Medical Center Utrecht. As of September 2004, he joined the Department of Information & Communication Technologies of the Universitat Pompeu Fabra in Barcelona, Spain, where he is currently an associate professor. He leads the Center for Computational Imaging & Simulation Technologies in Biomedicine. He is a senior member of the IEEE. More details about his research can be found at http://www.cilab.upf.edu.

**Lale Akarun** received the PhD degree in electronic engineering from the Polytechnic University, Brooklyn, New York, in 1992. Since 2002, she has been a professor in the Department of Computer Engineering at Bogazici University, Turkey. She is a senior member of the IEEE. More details about her research can be found at http://www.cmpe.boun.edu.tr/pilab.

**Arman Savran** recieved the MSc degree in electrical and electronics engineering in 2004 from Bogazici University, Turkey, where he is currently working toward the PhD degree in the Electrical and Electronics Engineering Department. More details about his research can be found at http://www.busim.ee.boun.edu.tr.